\documentclass[runningheads]{llncs}
\usepackage{graphicx}

\usepackage{tikz}
\usepackage{comment}
\usepackage{amsmath,amssymb} 
\usepackage{color}
\usepackage{booktabs}
\usepackage{textcomp}
\usepackage{multirow}
\usepackage{subcaption}

\begin{document}

\newcommand\todo[1]{\textcolor{blue}{#1}}

\pagestyle{headings}
\mainmatter
\def\ECCVSubNumber{100}  

\title{What Does CNN Shift Invariance Look Like?\\A Visualization Study}



\titlerunning{What Does CNN Shift Invariance Look Like? A Visualization Study}
%
\author{Jake Lee\inst{1} \and
Junfeng Yang\inst{1} \and
Zhangyang Wang\inst{2}}
\authorrunning{J. Lee et al.}
%
\institute{Columbia University, New York NY 10027, USA \\
\email{jake.h.lee@jpl.nasa.gov, junfeng@cs.columbia.edu} \and
The University of Texas at Austin, TX 78712, USA \\
\email{atlaswang@utexas.edu}}

\maketitle

\begin{abstract}

Feature extraction with convolutional neural networks (CNNs) is a popular method to represent images for machine learning tasks. These representations seek to capture global image content, and ideally should be independent of geometric transformations. We focus on measuring and visualizing the shift invariance of extracted features from popular off-the-shelf CNN models. We present the results of three experiments comparing representations of millions of images with exhaustively shifted objects, examining both local invariance (within a few pixels) and global invariance (across the image frame). We conclude that features extracted from popular networks are not globally invariant, and that biases and artifacts exist within this variance. Additionally, we determine that anti-aliased models significantly improve local invariance but do not impact global invariance. Finally, we provide a code repository for experiment reproduction, as well as a website to interact with our results at \url{https://jakehlee.github.io/visualize-invariance}.
\end{abstract}

\begin{figure}[h!]
    \centering
    \begin{subfigure}[b]{0.30\textwidth}
        \centering
        \includegraphics[trim={2cm 0 2cm 0}, clip, width=\textwidth]{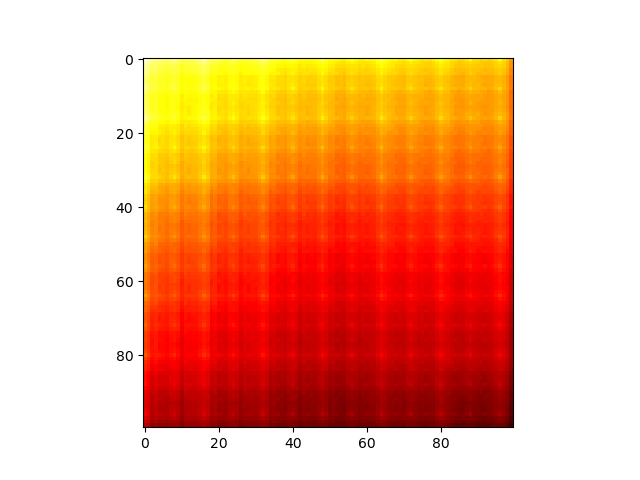}
        \caption{\centering AlexNet fc8 \newline off-the-shelf}
        \label{subfig:intro-a}
    \end{subfigure}
    \begin{subfigure}[b]{0.30\textwidth}
        \centering
        \includegraphics[trim={2cm 0 2cm 0}, clip, width=\textwidth]{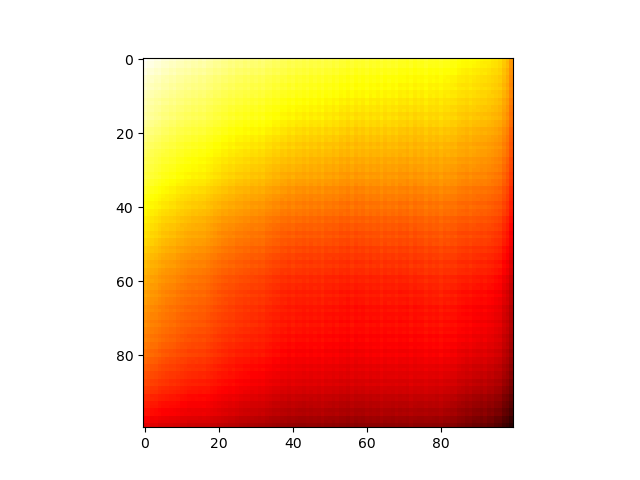}
        \caption{\centering AlexNet fc8 \newline anti-aliased \cite{zhang:shiftinv}}
        \label{subfig:intro-aa}
    \end{subfigure}
    \caption{An example heatmap of cosine similarities as an indicator for shift invariance, using features extracted from the last fully-connected layer of AlexNet. Features from each shift location are compared to the features of an image with the object at the top left. Brighter colors at each shift location indicate more similar features. Refer to Section \ref{sec:res-sim} for more results.}
    \label{fig:intro-heatmap}
\end{figure}

\section{Introduction}

Convolutional neural networks (CNNs) are able to achieve state-of-the-art performance on computer vision tasks such as object classification \cite{krizhevsky:imagenet,he:resnet} and image segmentation \cite{he:maskrcnn}. Transfer learning methods \cite{yosinski:transfer} allow tasks to leverage models pre-trained on large, generic datasets such as ImageNet \cite{deng:imagenet} and MIT Places \cite{zhou:places} instead of training a CNN from scratch, which can be costly. One popular method is extracting neuron activations of a layer in the pre-trained CNN and treating them as feature representations of the input image \cite{sharif:astounding}. Using these feature vectors with machine learning methods result in competitive performance for classification in different domains \cite{sharif:astounding,schwarz:rgbdpre}, content based image retrieval \cite{babenko:aggregating}, and novel image detection \cite{lee:visualizing}.

However, it is known that CNNs, and therefore CNN features, lack geometric invariance. Simple, small transformations such as translations and rotations can significantly impact classification accuracy \cite{engstrom:rotation,azulay:whypoor,pei2017towards}. For feature extraction, geometric invariance may be desired to retrieval a global descriptor of the image robust to minor changes in image capture. This goal is especially relevant for content based image retrieval tasks, as images only a small shift or rotation apart should result in similar features \cite{babenko:aggregating}. However, any task that relies on CNN feature extraction would benefit from models more robust to geometric transformations, as the extracted features would better represent the content of the images.

Several methods to improve geometric invariance have been proposed, including geometric training dataset augmentation \cite{shorten:augmentation}, spatial transformer networks \cite{jaderberg:spatialtransf}, and anti-aliasing \cite{zhang:shiftinv}. For improving the invariance of extracted features, methods have been proposed in the context of image retrieval \cite{gong:multiscale} by post-processing the extracted features instead of making the model itself more robust. Despite these proposals, it remains questionable to what extent we can trust state-of-the-art classifier networks as translation-invariant, even with those add-ons, and there is a lack of principled study on examining that either qualitatively or quantitatively.

In this work, we examine the shift invariance of features extracted from off-the-shelf pre-trained CNNs with visualizations and quantitative experiments. These experiments take advantage of exhaustive testing of the input space, extracting features from millions of images with objects at different locations in the image frame. Such fine-grained experiments allow for observing invariance from \textit{two different lenses}: both \textbf{global shift invariance}, defined the the maximal coverage of object location translations that the feature can stay relatively invariant to (based some feature similarity threshold); and \textbf{local shift invariance}, defined as the relative invariance changes when perturbing the object location for just a few pixels. Note that most literature discussed the translation invariance through the latter lens \cite{shorten:augmentation,zhang:shiftinv}, while we advocate both together for more holistic and fine-grained understanding.

\begin{figure}[t]
    \centering
    \includegraphics[width=\textwidth]{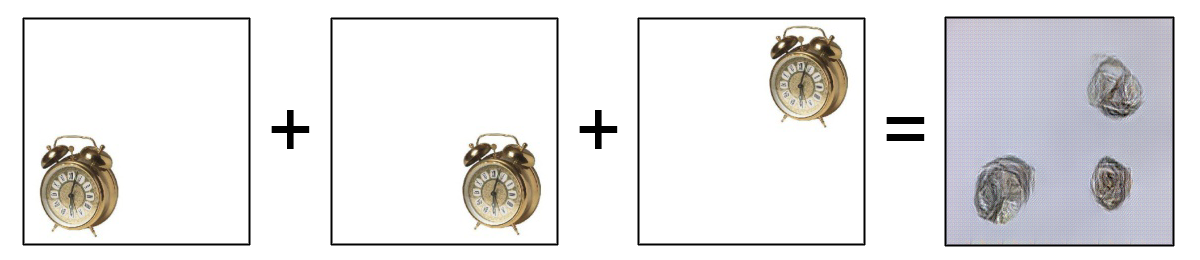}
    \caption{``Feature arithmetic" performed with features extracted from AlexNet's fc7 layer. Features were extracted from operand images, added together, and visualized with DeepSiM \cite{dosovitskiy:deepsim}. That provides empirical evidence that the extracted fully-connected features still preserve almost complete spatial information, which could be considered to ``counterexamples" to the claimed shift invariance of popular deep models. Refer to Section \ref{sec:res-arith} for complete results.}
    \label{fig:intro-arith}
\end{figure}

We focus on features extracted from fully-connected layers of the models, as they are often assumed to be more geometrically invariant \cite{lench:rcnnr} than convolutional feature maps. We also compare the robustness of standard pre-trained models \cite{chen2020adversarial}, and those models trained by a state-of-the-art anti-aliasing technique to boost translation invariance \cite{zhang:shiftinv}. We draw the following main observations:
\begin{itemize}
    \item Visualizing cosine similarities of features from shifted images show that almost all existing pre-trained classifiers' extracted features, even from high-level fully-connected layers, are brittle to input translations. Those most accurate models, such as ResNet-50, suffer even more from translation fragility than simpler ones such as AlexNet.
    \item Interestingly, we observe an empirical bias towards greater similarity for horizontally shifted images, compared to vertical translations. Also, a grid pattern was universally observed in vanilla pre-trained models, showing significant local invariance fluctuations and concurring the observation in \cite{zhang:shiftinv}. An example is shown in Figure \ref{subfig:intro-a}.
    \item Antialiased models \cite{zhang:shiftinv} are able to suppress such local fluctuations and improve local invariance, but do not visibly improve the global invariance. An example is shown in Figure \ref{subfig:intro-aa}.
    \item A side product that we create is a ``feature arithmetic" demonstration: adding or subtracting features extracted from images with shifted object locations, then visualizing the result hidden feature to the pixel domain. That results in images with objects added or removed spatially correctly, despite the lack of any latent space optimization towards that goal. An example is shown in \ref{fig:intro-arith}, which we suggest may serve as another ``counterexamples" to the claimed shift invariance of popular deep models.
\end{itemize}

\section{Methods}

We perform three sets of experiments to describe and quantify the robustness of extracted features to object translation. First, we measure the sensitivity of extracted features to translation by calculating cosine similarities between features of translated patches (Section \ref{sec:feat_sim}). Next, we train a linear SVM on extracted features of translated patches to measure basic separability (Section \ref{sec:feat_sep}). Finally, we demonstrate that extracted features can be added and subtracted, and coherent spatial information can still be recovered (Section \ref{sec:feat_arith}).

\subsection{Feature Similarity} \label{sec:feat_sim}

We adopt Zhang's definition of shift invariance \cite{zhang:shiftinv}: an extracted feature is shift invariant if shifting the input image results in an identical extracted feature. To quantify the invariance, we exhaustively test the input space. We define a segmented object image patch and generate all translations of the patch across a white background with a stride of one pixel. Compared to an alternative method---exhaustively cropping a smaller image from a larger image (similarly to data augmentation to prevent overfitting \cite{krizhevsky:imagenet}), our method ensures that the pixel content is exactly the same for every image and the only variation is the geometric transformation.

After feature extraction, we calculate the cosine similarity between vectors. Since similarity is comparative, we define five anchor patch locations to compare vectors against: top left, top right, center, bottom left, and bottom right. By our definition, a completely shift-invariant model would have identical extracted features for every image, and a cosine similarity of 1 for every comparison. However, a less shift-invariant model would have lower cosine similarities. By observing the results for each anchor, we can determine if the features are more sensitive to shifts in a certain axis, or if certain corners of the image are more sensitive than others.

\subsection{Feature Separability} \label{sec:feat_sep}

To evaluate the shift invariance of features beyond consine similarity, we also measure the separability of the extracted features. While the previous experiments quantify feature similarity, it does not show whether spatial information is well-encoded by the features. It is possible that the features are not shift invariant, but the differences due to shifting are random. It is also possible that the changes are well-correlated with patch location, and therefore separable.

To measure this separability, we train a linear SVM classifier on the extracted features to determine whether the patch is on the top or bottom half of the image, and another to determine whether the patch is on the left or right half of the image. We perform 5-fold stratified cross-validation for each patch object. 

A completely shift invariant model would generate identical features for every image, resulting in random classifier accuracy (50\%). A less shift invariant model may still generate features not correlated with the patch location. In this case, we can still expect near-random classifier accuracy. Finally, a less shift invariant model may generate features well-correlated with the patch location, resulting in higher classifier accuracy.

\subsection{Feature Arithmetic} \label{sec:feat_arith}

Finally, we consider a case in which extracted features have encoded shift information and spatial information very well. It may be possible to manipulate extracted features and still recover information, similarly to latent space interpolation or arithmetic in autoencoders and generative adversarial networks (GANs) \cite{bojanowski:latent,Radford:gans}. Dosovitskiy and Brox have previously shown that images can be recovered from features extracted from layers of AlexNet, as well as from interpolated features \cite{dosovitskiy:deepsim}. Performing feature arithmetic with these features may give us further insight into the shift invariance of extracted features.

We perform feature arithmetic by extracting features of images with objects in different locations. We then add or subtract the extracted features, hoping to add or remove images of said objects in different locations. Finally, we visualize these features with the GAN model proposed by Dosovitskiy and Brox (referred to as DeePSiM) \cite{dosovitskiy:deepsim}. We only perform these experiments with features extracted from the fc6, fc7, and fc8 layers of AlexNet \cite{krizhevsky:imagenet}, as DeePSiM architectures were only provided for those layers. However, we make a slight adjustment and use features extracted prior to ReLU activation to take advantage of the additional information. We used re-trained DeePSim model weights provided by Lee and Wagstaff \cite{lee:visualizing} for this modification.

If the extracted features encode location information well, then the visualizations of the results of arithmetic should be similar to what we would have expected had we performed the arithmetic in pixel space. It should be noted that, in contrast to unsupervised GANs trained to learn an efficient latent space that can be recovered \cite{Radford:gans}, AlexNet had no motivation to learn such a latent space when training for ImageNet object classification. Therefore, it is already impressive that the input image can be successfully recovered from features extracted from the fully connected layers. If meaningful images can be recovered from added or subtracted features, it would further show that the features encode spatial information in addition to class and object information.

\section{Results}

\subsection{Datasets}

For our feature similarity and separability experiments, we use 20 segmented objects each from 10 classes in the COCO 2017 training set as patches \cite{lin:coco}: \texttt{car, airplane, boat, dog, bird, zebra, orange, banana, clock,} and \texttt{laptop}. These classes were selected because of their range of object types, and because the same classes also exist in ImageNet \cite{deng:imagenet}, on which all of our models being evaluated are trained. This does not mean that the results of the experiment are exclusive to ImageNet, however; feature extraction is known to be applicable domains outside of the training set \cite{sharif:astounding}.

The white background images are 224x224 in size. We resize each segmented and cropped object to 75x75 (small) and 125x125 (large) to explore translating patches of different sizes. Each patch is then translated with stride 1 to every possible location in the 224x224 frame. This results in 22.5k images for the smaller patch and 10k images for the larger patch. In total, for 200 unique patches, there are 4.5 million small-patch and 2 million large-patch images.

For our feature arithmetic experiments, we use a handcrafted dataset of object image patches against a white background. We place some of the same image patches at specific locations so that, when added or subtracted, multiple patches can appear or patches can be removed. These images are described in further detail in Section \ref{sec:res-arith}. 

\subsection{Feature Extraction Models} \label{sec:models}

For our feature similarity and separability experiments, we evaluate the following popular pre-trained models provided by Pytorch \cite{paszke:pytorch}: AlexNet \cite{krizhevsky:imagenet}, ResNet-50 \cite{he:resnet}, and MobileNetV2 \cite{sandler:mobilenet}. 

We extract features from all three fully-connected layers of AlexNet to determine if and how shift invariance changes deeper into the network. ResNet-50 has only one fully-connected layer (the final classification layer prior to softmax activation), but is far deeper overall than AlexNet, enabled by its skip connections. It also has a global average pooling layer prior to the final fully connected layer. Finally, MobileNetV2 also only has one fully-connected layer as the final classification layer prior to softmax activation.

We also evaluate the anti-aliased versions of these models by Zhang \cite{zhang:shiftinv}, as they claim to improve shift invariance and classification accuracy. We use the Bin-5 filter, as they reported the best ``consistency'' compared to other filters.

All models were pre-trained on ILSVRC2012 \cite{deng:imagenet} with data augmentation.\footnote{\url{https://github.com/pytorch/vision/blob/master/references/classification/train.py\\\#L96-L103}} First, a crop of random size (from 0.08 to 1.0 of the original size) and random aspect ratio (3/4 to 4/3) is made, and is resized to 224x224. Second, the image is flipped horizontally with 0.5 probability. For each model, we extract features from fully connected layers prior to ReLU or Softmax activation.

\subsection{Feature Similarity} \label{sec:res-sim}

\begin{figure}[p]
    \centering
    \begin{subfigure}[b]{0.23\textwidth}
        \centering
        \includegraphics[trim={2cm 0 2cm 0}, clip, width=\textwidth]{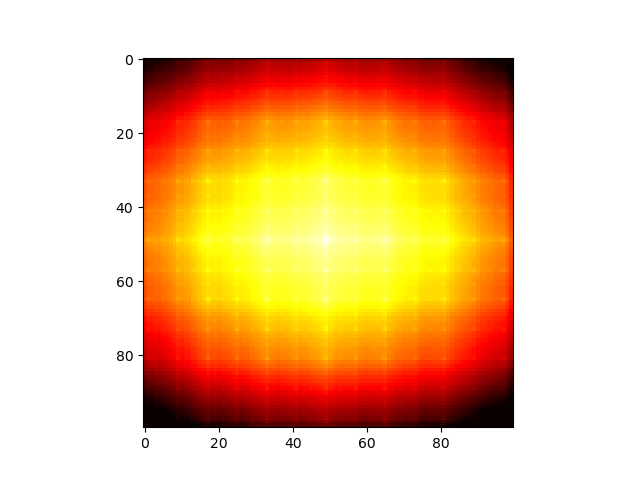}
        \caption{Alex fc6 C} \label{fig:3a}
    \end{subfigure}
    \hfill
    \begin{subfigure}[b]{0.23\textwidth}
        \centering
        \includegraphics[trim={2cm 0 2cm 0}, clip, width=\textwidth]{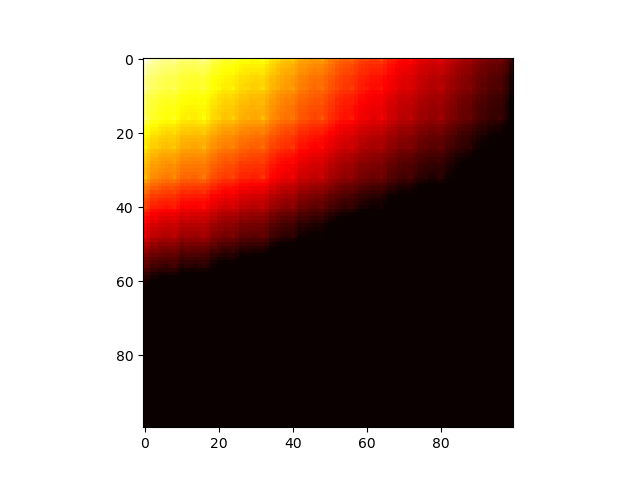}
        \caption{Alex fc6 TL}
    \end{subfigure}
    \hfill
    \begin{subfigure}[b]{0.23\textwidth}
        \centering
        \includegraphics[trim={2cm 0 2cm 0}, clip, width=\textwidth]{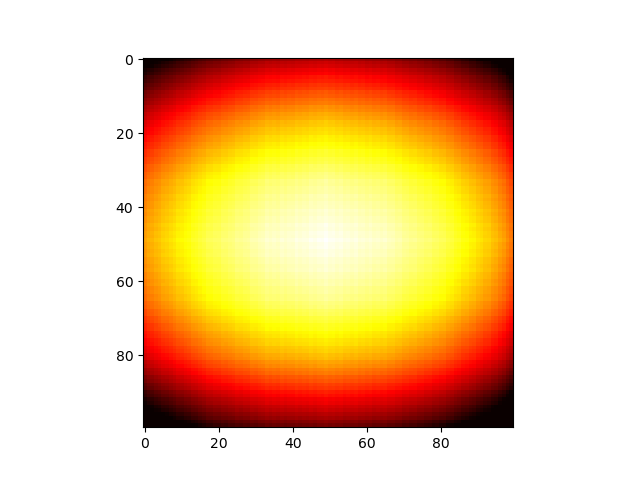}
        \caption{Alex-AA fc6 C}
    \end{subfigure}
    \hfill
    \begin{subfigure}[b]{0.23\textwidth}
        \centering
        \includegraphics[trim={2cm 0 2cm 0}, clip, width=\textwidth]{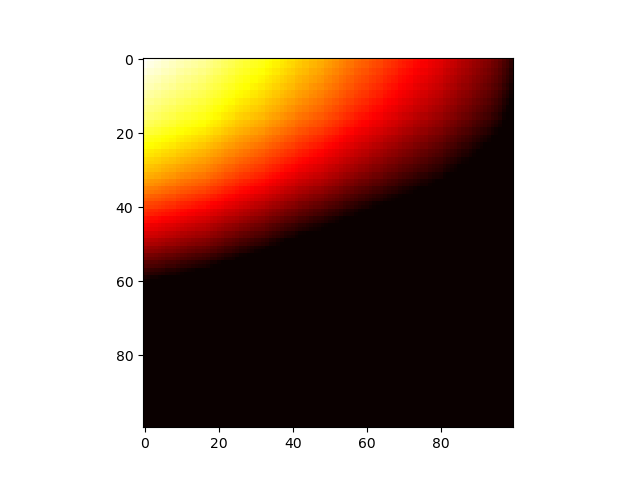}
        \caption{Alex-AA fc6 TL}
    \end{subfigure}
    
    \begin{subfigure}[b]{0.23\textwidth}
        \centering
        \includegraphics[trim={2cm 0 2cm 0}, clip, width=\textwidth]{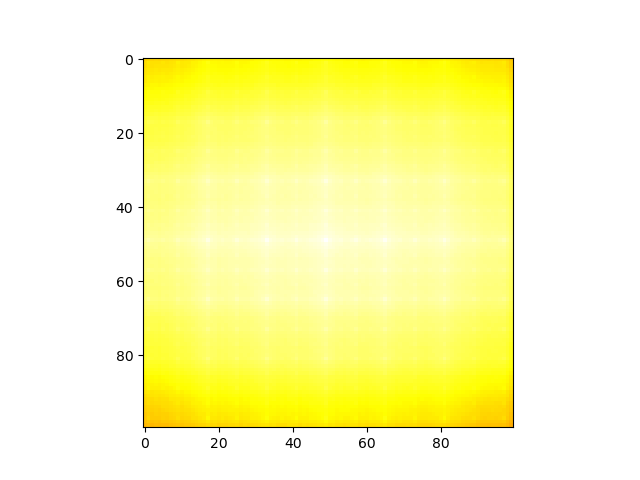}
        \caption{Alex fc7 C}
    \end{subfigure}
    \hfill
    \begin{subfigure}[b]{0.23\textwidth}
        \centering
        \includegraphics[trim={2cm 0 2cm 0}, clip, width=\textwidth]{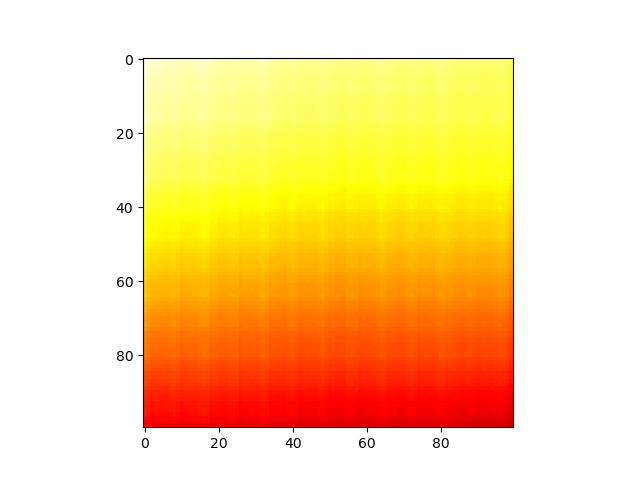}
        \caption{Alex fc7 TL} \label{fig:3f}
    \end{subfigure}
    \hfill
    \begin{subfigure}[b]{0.23\textwidth}
        \centering
        \includegraphics[trim={2cm 0 2cm 0}, clip, width=\textwidth]{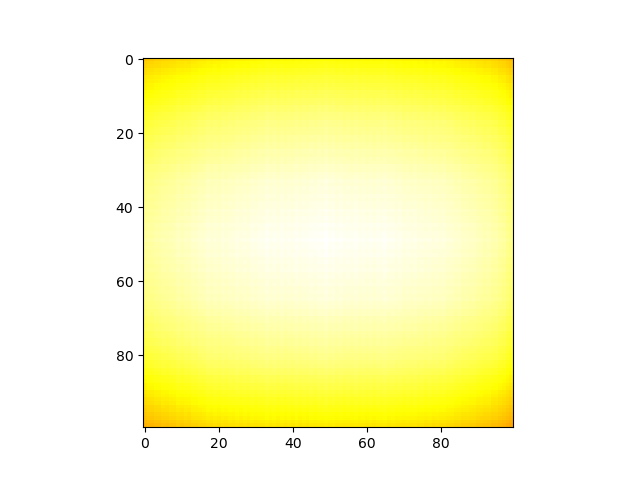}
        \caption{Alex-AA fc7 C}
    \end{subfigure}
    \hfill
    \begin{subfigure}[b]{0.23\textwidth}
        \centering
        \includegraphics[trim={2cm 0 2cm 0}, clip, width=\textwidth]{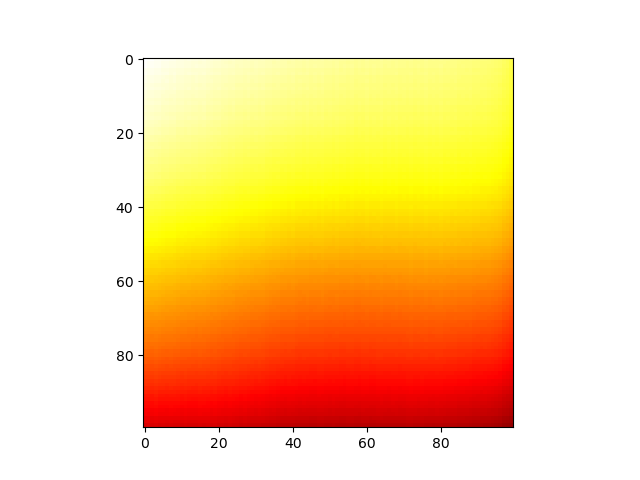}
        \caption{Alex-AA fc7 TL}
    \end{subfigure}
    
    \begin{subfigure}[b]{0.23\textwidth}
        \centering
        \includegraphics[trim={2cm 0 2cm 0}, clip, width=\textwidth]{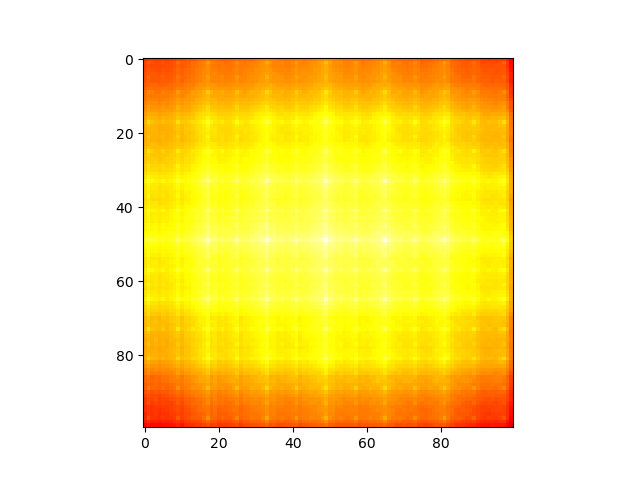}
        \caption{Alex fc8 C}
    \end{subfigure}
    \hfill
    \begin{subfigure}[b]{0.23\textwidth}
        \centering
        \includegraphics[trim={2cm 0 2cm 0}, clip, width=\textwidth]{alexnetfc8/large-tl.png}
        \caption{Alex fc8 TL}
    \end{subfigure}
    \hfill
    \begin{subfigure}[b]{0.23\textwidth}
        \centering
        \includegraphics[trim={2cm 0 2cm 0}, clip, width=\textwidth]{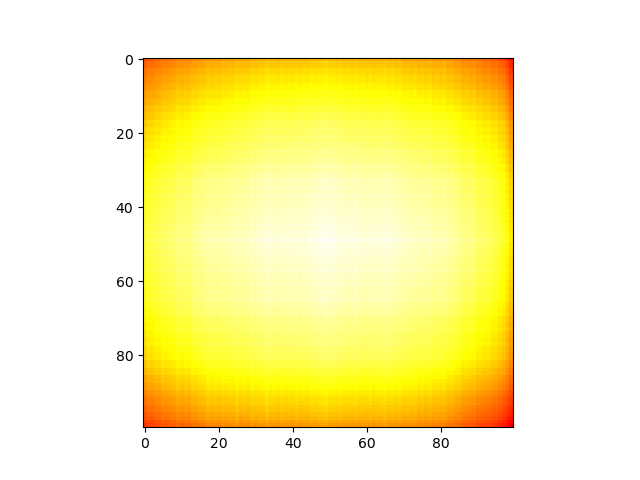}
        \caption{Alex-AA fc8 C}
    \end{subfigure}
    \hfill
    \begin{subfigure}[b]{0.23\textwidth}
        \centering
        \includegraphics[trim={2cm 0 2cm 0}, clip, width=\textwidth]{alexnetfc8/large-AA-tl.png}
        \caption{Alex-AA fc8 TL}
    \end{subfigure}
    
    \begin{subfigure}[b]{0.23\textwidth}
        \centering
        \includegraphics[trim={2cm 0 2cm 0}, clip, width=\textwidth]{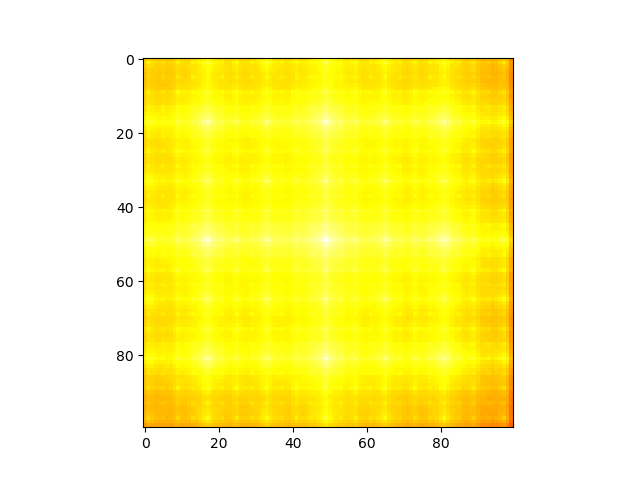}
        \caption{ResN50 C}
    \end{subfigure}
    \hfill
    \begin{subfigure}[b]{0.23\textwidth}
        \centering
        \includegraphics[trim={2cm 0 2cm 0}, clip, width=\textwidth]{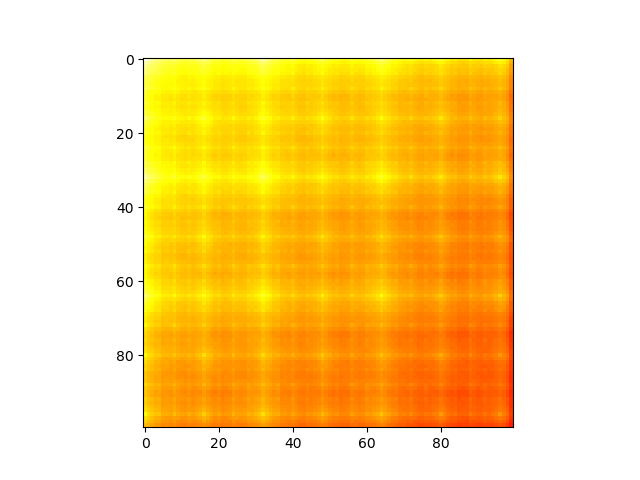}
        \caption{ResN50 TL}
    \end{subfigure}
    \hfill
    \begin{subfigure}[b]{0.23\textwidth}
        \centering
        \includegraphics[trim={2cm 0 2cm 0}, clip, width=\textwidth]{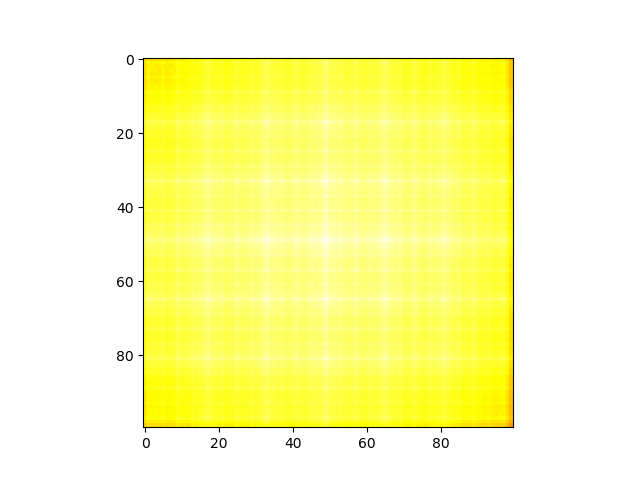}
        \caption{ResN50-AA C}
    \end{subfigure}
    \hfill
    \begin{subfigure}[b]{0.23\textwidth}
        \centering
        \includegraphics[trim={2cm 0 2cm 0}, clip, width=\textwidth]{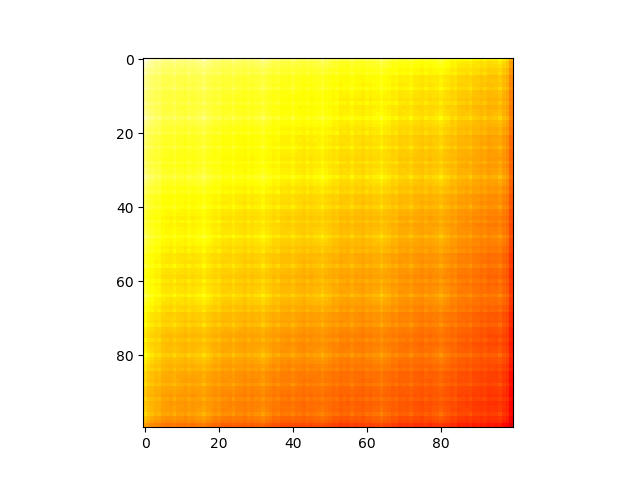}
        \caption{ResN50-AA TL}
    \end{subfigure}
    
    \begin{subfigure}[b]{0.23\textwidth}
        \centering
        \includegraphics[trim={2cm 0 2cm 0}, clip, width=\textwidth]{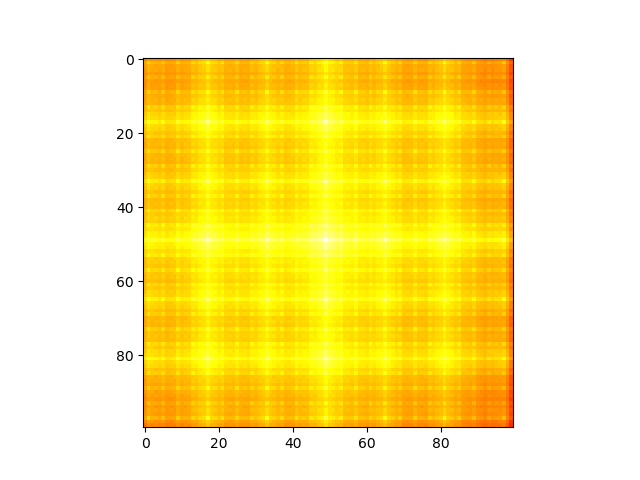}
        \caption{Mobile C}
    \end{subfigure}
    \hfill
    \begin{subfigure}[b]{0.23\textwidth}
        \centering
        \includegraphics[trim={2cm 0 2cm 0}, clip, width=\textwidth]{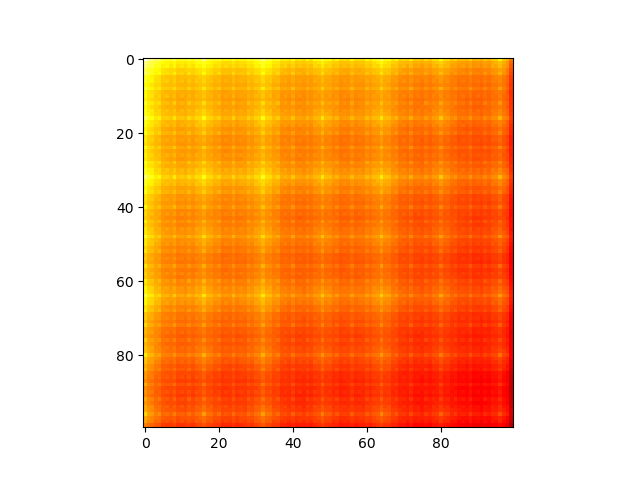}
        \caption{Mobile TL}
    \end{subfigure}
    \hfill
    \begin{subfigure}[b]{0.23\textwidth}
        \centering
        \includegraphics[trim={2cm 0 2cm 0}, clip, width=\textwidth]{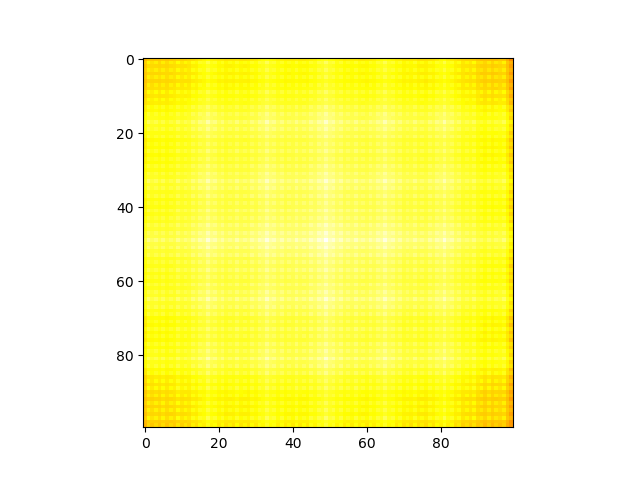}
        \caption{Mobile-AA C}
    \end{subfigure}
    \hfill
    \begin{subfigure}[b]{0.23\textwidth}
        \centering
        \includegraphics[trim={2cm 0 2cm 0}, clip, width=\textwidth]{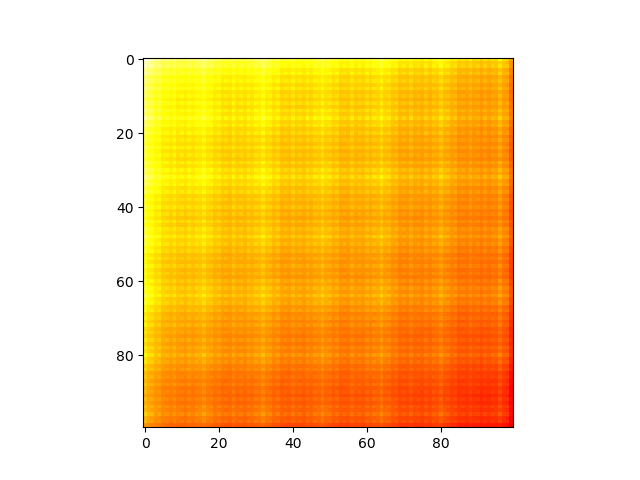}
        \caption{Mobile-AA TL}
    \end{subfigure}
    
    \begin{subfigure}[b]{0.8\textwidth}
        \centering
        \includegraphics[width=\textwidth]{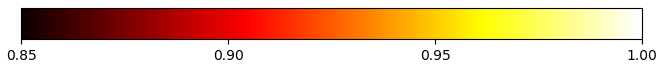}
    \end{subfigure}
    \caption{Cosine similarity heatmaps of the large-patch dataset. Each pixel represents the average cosine similarity of different object patches at that shift location. C indicates the center anchor and TL indicates the top-left anchor.}
    \label{fig:simlarge}
\end{figure}

First, observe the similarities of features extracted from the large-patch dataset. Figure \ref{fig:simlarge} shows heatmaps visualizing the average cosine similarities of 200 object patches at each shift location. Only comparisons with the center (C) and top-left (TL) anchor points are shown. Heatmaps for all comparison anchor points can be seen at \url{https://jakehlee.github.io/visualize-invariance}. 

Overall, features extracted from the fc7 layer of AlexNet seem to be the most shift invariant, with the anti-aliased ResNet-50 close behind. This is also supported by Table \ref{tab:simlarge}, which reports the mean and standard deviation of the mean cosine similarity for each object patch.

Interestingly, while similarities with the center anchor remain fairly consistent (with lower similarities at the corners), results for the top-left anchor show that extracted features remain more similar when the patch is shifted horizontally than vertically. This is most clearly seen in Figure \ref{fig:3f}. Additionally, the heatmap in Figure \ref{fig:3a} takes the form of an oval stretched horizontally, indicating the same bias. We hypothesized that this may be due to random horizontal flipping included in the training data augmentation. The model may have encountered more objects varying horizontally than vertically during training, resulting in greater invariance along that axis. 

\begin{figure}[t]
    \centering
    \begin{subfigure}[b]{0.23\textwidth}
        \centering
        \includegraphics[trim={2cm 0 2cm 0}, clip, width=\textwidth]{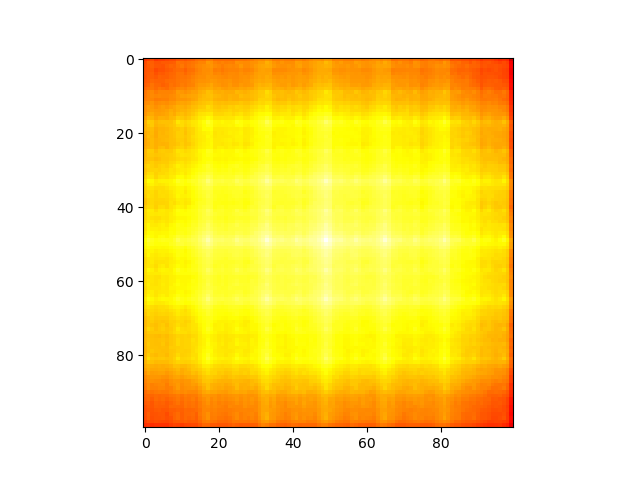}
        \caption{Center \newline with Horiz. Flip}
        \label{subfig:flip-c}
    \end{subfigure}
    \begin{subfigure}[b]{0.23\textwidth}
        \centering
        \includegraphics[trim={2cm 0 2cm 0}, clip, width=\textwidth]{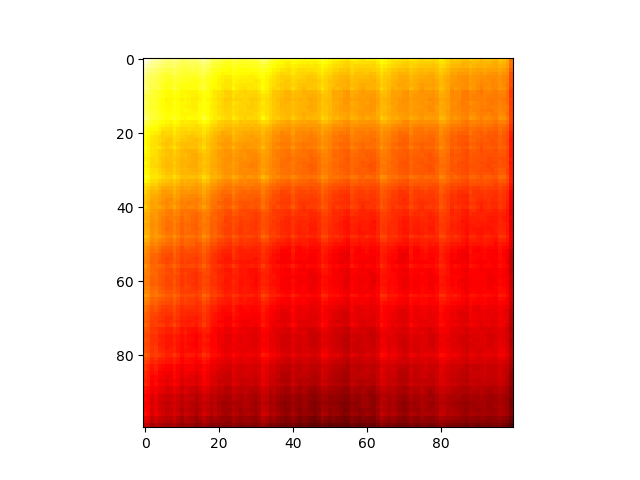}
        \caption{Top-Left \newline with Horiz. Flip}
        \label{subfig:flip-tl}
    \end{subfigure}
    \begin{subfigure}[b]{0.23\textwidth}
        \centering
        \includegraphics[trim={2cm 0 2cm 0}, clip, width=\textwidth]{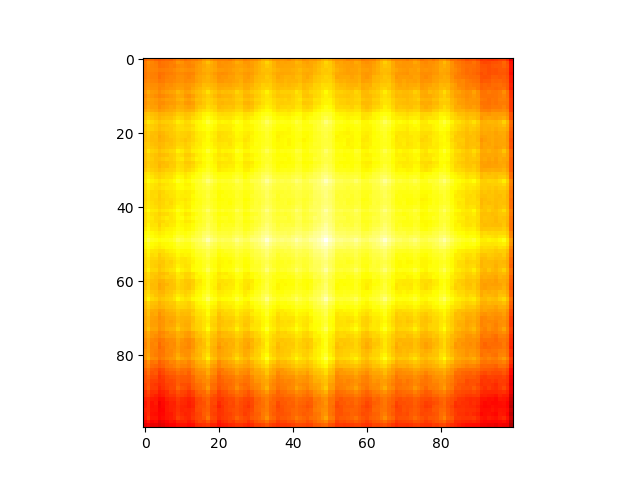}
        \caption{Center \newline w/o Horiz. Flip}
        \label{subfig:noflip-c}
    \end{subfigure}
    \begin{subfigure}[b]{0.23\textwidth}
        \centering
        \includegraphics[trim={2cm 0 2cm 0}, clip, width=\textwidth]{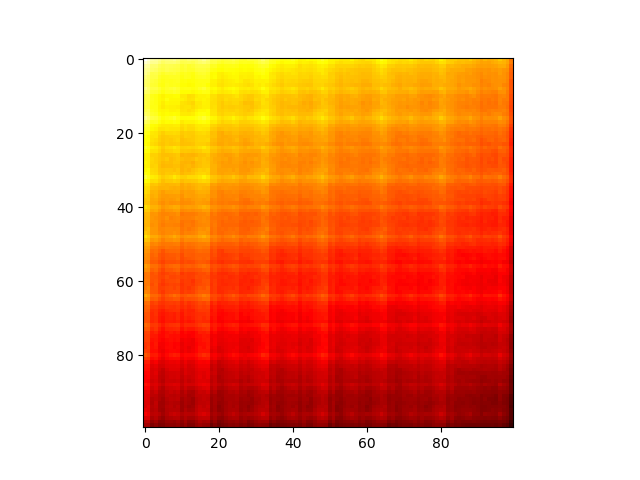}
        \caption{Top-Left \newline w/o Horiz. Flip}
        \label{subfig:noflip-tl}
    \end{subfigure}
    \caption{Cosine similarity heatmaps of AlexNet fc8 features on the large-patch dataset. Models were trained with and without random horizontal flip augmentation for 5 epochs.}
    \label{fig:flip-exp}
\end{figure}

To test this hypothesis, we trained AlexNet models on the ILSVRC2012 ImageNet training set \cite{ILSVRC15} with and without random horizontal flip augmentation for 5 epochs using PyTorch-provided training hyperparameters.\footnote{https://github.com/pytorch/vision/tree/master/references/classification} While the original models were trained for 90 epochs, we believe 5 epochs are sufficient for comparison. Heatmaps generated with these two models are shown in Figure \ref{fig:flip-exp}. The heatmaps reveal no significant differences between the two models, rejecting our hypothesis. The bias must be a result of the dataset or architecture, not the training set augmentation.

Additionally, we can observe the general improvement that anti-aliasing provides to shift invariance and overall consistency. In the heatmaps for all layers of off-the-shelf models, there is a clear grid pattern visible, most distinct in AlexNet's fc8, ResNet-50, and MobileNetV2. The anti-aliased models either completely eliminate or significantly suppress this grid pattern, improving the consistency and local invariance. Additionally, for some layers, the antialiased model also slightly improves overall similarity and global invariance.

\begin{figure}[p]
    \centering
    \begin{subfigure}[b]{0.23\textwidth}
        \centering
        \includegraphics[trim={2cm 0 2cm 0}, clip, width=\textwidth]{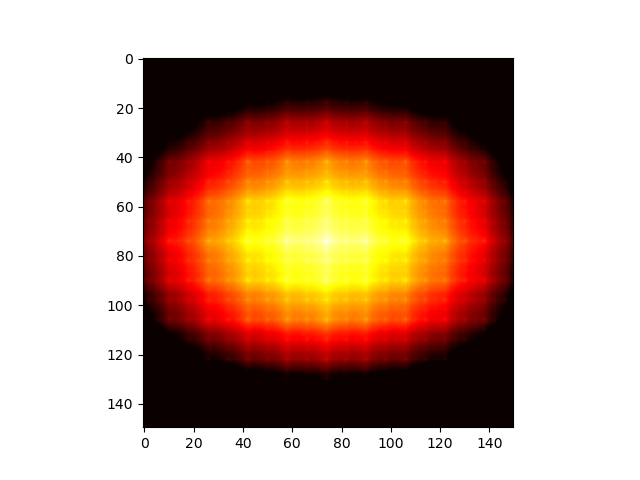}
        \caption{Alex fc6 C}
    \end{subfigure}
    \hfill
    \begin{subfigure}[b]{0.23\textwidth}
        \centering
        \includegraphics[trim={2cm 0 2cm 0}, clip, width=\textwidth]{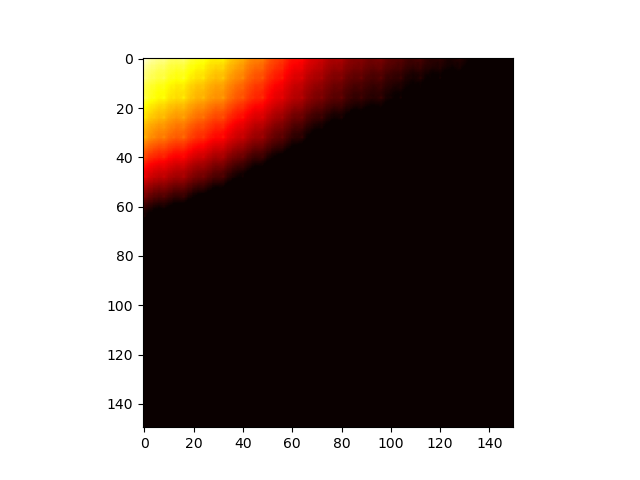}
        \caption{Alex fc6 TL}
    \end{subfigure}
    \hfill
    \begin{subfigure}[b]{0.23\textwidth}
        \centering
        \includegraphics[trim={2cm 0 2cm 0}, clip, width=\textwidth]{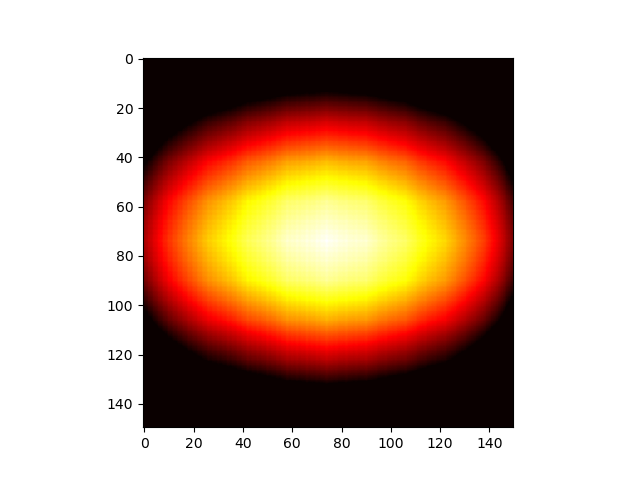}
        \caption{Alex-AA fc6 C}
    \end{subfigure}
    \hfill
    \begin{subfigure}[b]{0.23\textwidth}
        \centering
        \includegraphics[trim={2cm 0 2cm 0}, clip, width=\textwidth]{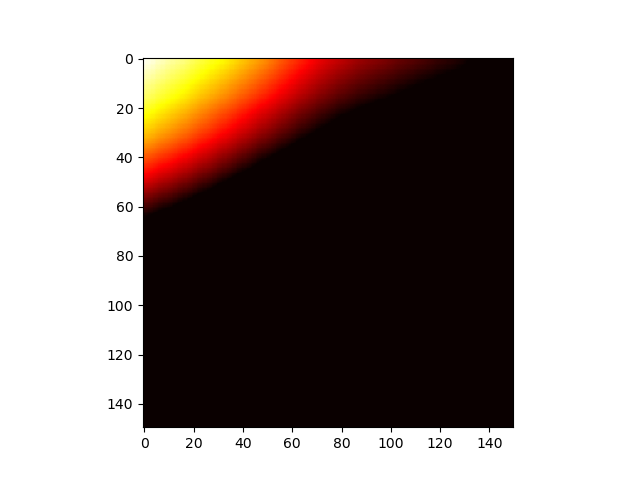}
        \caption{Alex-AA fc6 TL}
    \end{subfigure}
    
    \begin{subfigure}[b]{0.23\textwidth}
        \centering
        \includegraphics[trim={2cm 0 2cm 0}, clip, width=\textwidth]{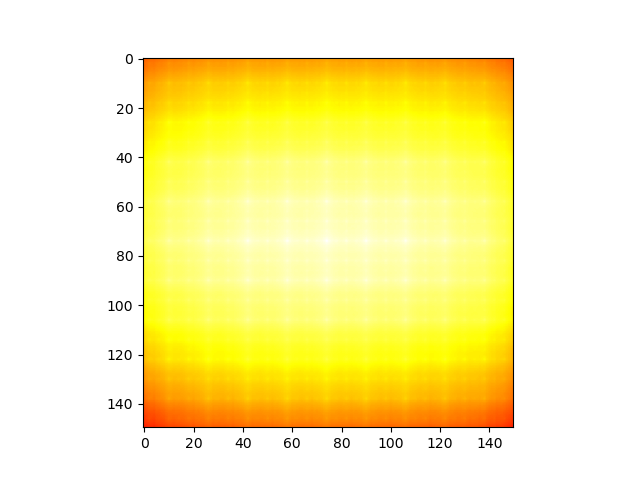}
        \caption{Alex fc7 C}
    \end{subfigure}
    \hfill
    \begin{subfigure}[b]{0.23\textwidth}
        \centering
        \includegraphics[trim={2cm 0 2cm 0}, clip, width=\textwidth]{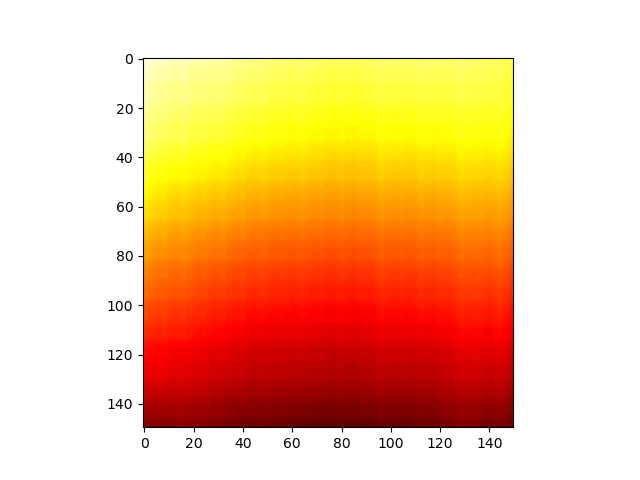}
        \caption{Alex fc7 TL}
    \end{subfigure}
    \hfill
    \begin{subfigure}[b]{0.23\textwidth}
        \centering
        \includegraphics[trim={2cm 0 2cm 0}, clip, width=\textwidth]{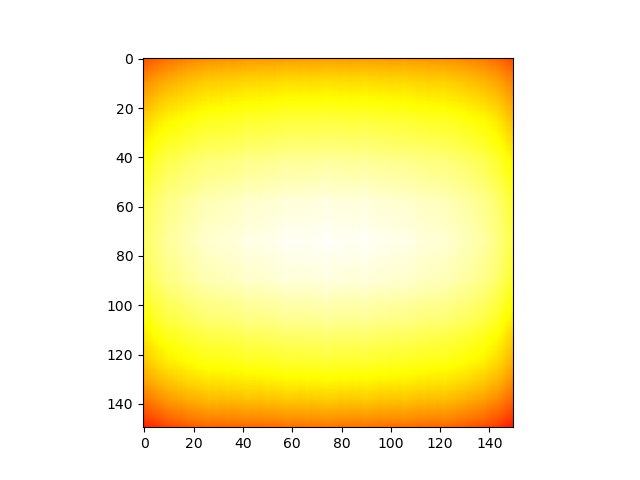}
        \caption{Alex-AA fc7 C}
    \end{subfigure}
    \hfill
    \begin{subfigure}[b]{0.23\textwidth}
        \centering
        \includegraphics[trim={2cm 0 2cm 0}, clip, width=\textwidth]{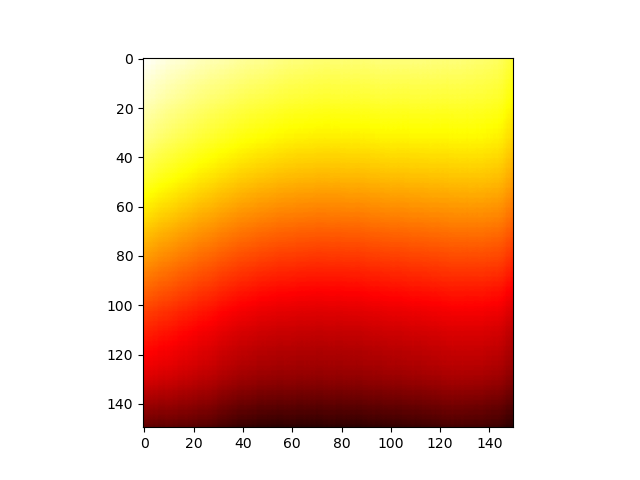}
        \caption{Alex-AA fc7 TL}
    \end{subfigure}
    
    \begin{subfigure}[b]{0.23\textwidth}
        \centering
        \includegraphics[trim={2cm 0 2cm 0}, clip, width=\textwidth]{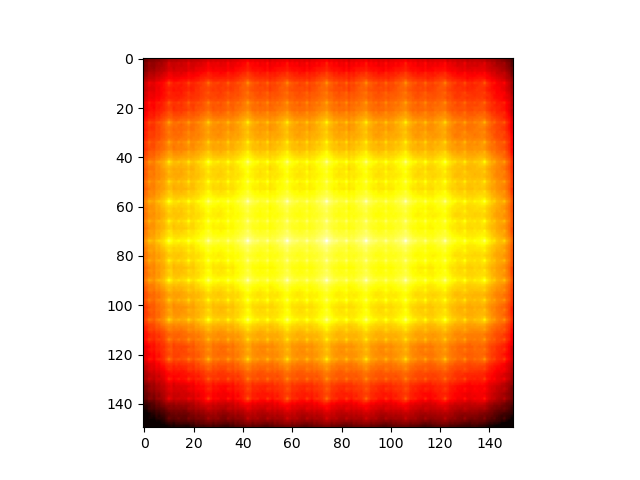}
        \caption{Alex fc8 C}
    \end{subfigure}
    \hfill
    \begin{subfigure}[b]{0.23\textwidth}
        \centering
        \includegraphics[trim={2cm 0 2cm 0}, clip, width=\textwidth]{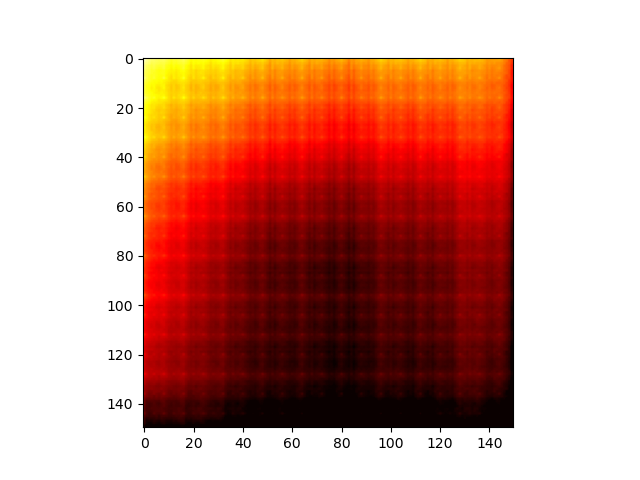}
        \caption{Alex fc8 TL}
    \end{subfigure}
    \hfill
    \begin{subfigure}[b]{0.23\textwidth}
        \centering
        \includegraphics[trim={2cm 0 2cm 0}, clip, width=\textwidth]{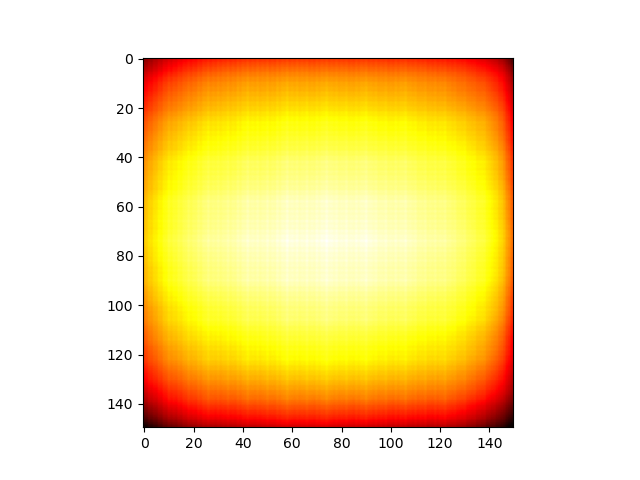}
        \caption{Alex-AA fc8 C}
    \end{subfigure}
    \hfill
    \begin{subfigure}[b]{0.23\textwidth}
        \centering
        \includegraphics[trim={2cm 0 2cm 0}, clip, width=\textwidth]{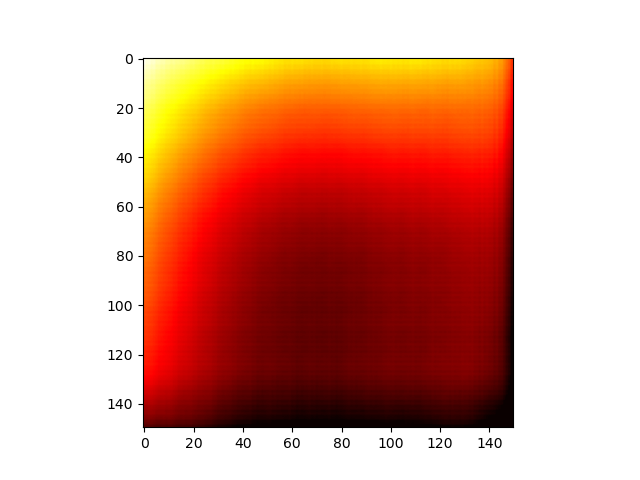}
        \caption{Alex-AA fc8 TL}
    \end{subfigure}
    
    \begin{subfigure}[b]{0.23\textwidth}
        \centering
        \includegraphics[trim={2cm 0 2cm 0}, clip, width=\textwidth]{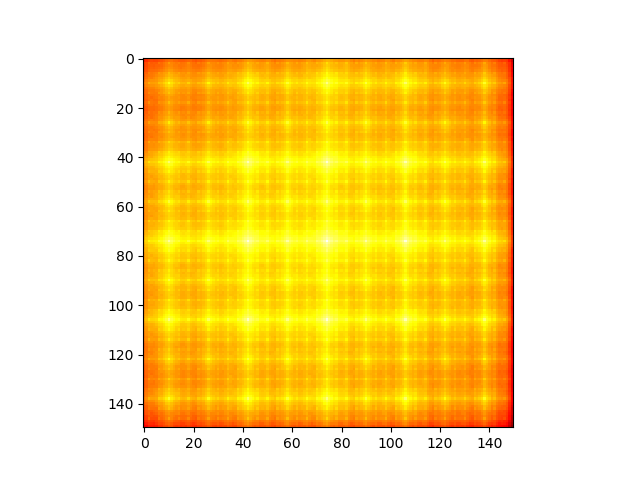}
        \caption{ResN50 C}
    \end{subfigure}
    \hfill
    \begin{subfigure}[b]{0.23\textwidth}
        \centering
        \includegraphics[trim={2cm 0 2cm 0}, clip, width=\textwidth]{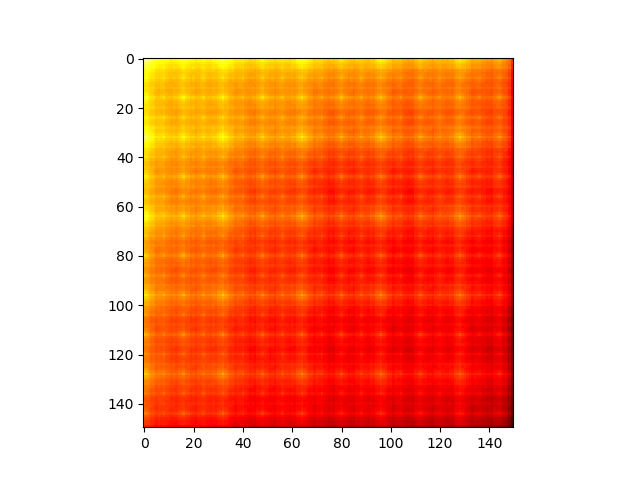}
        \caption{ResN50 TL}
    \end{subfigure}
    \hfill
    \begin{subfigure}[b]{0.23\textwidth}
        \centering
        \includegraphics[trim={2cm 0 2cm 0}, clip, width=\textwidth]{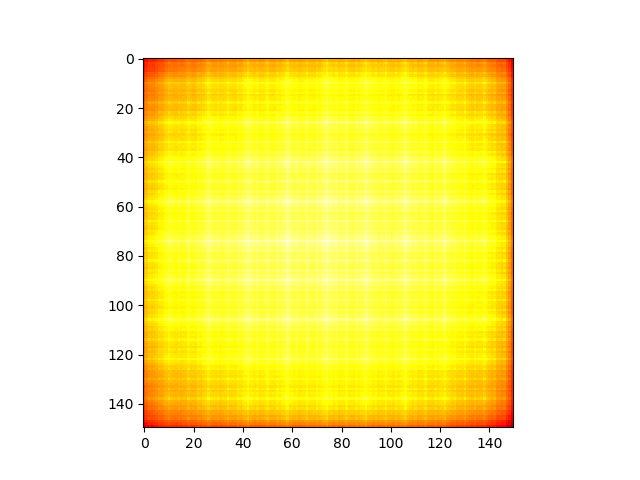}
        \caption{ResN50-AA C}
    \end{subfigure}
    \hfill
    \begin{subfigure}[b]{0.23\textwidth}
        \centering
        \includegraphics[trim={2cm 0 2cm 0}, clip, width=\textwidth]{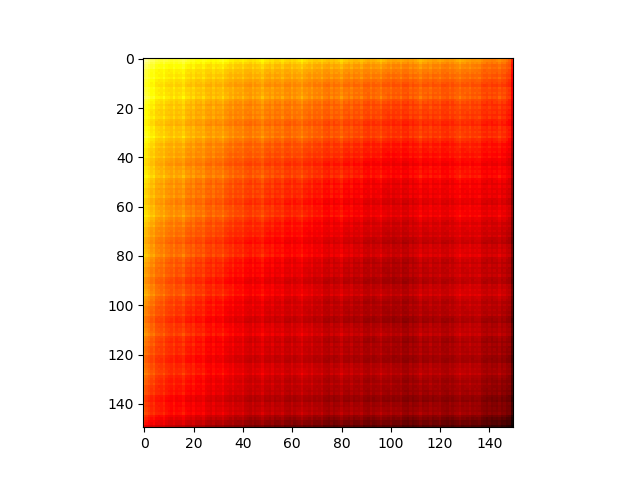}
        \caption{ResN50-AA TL}
    \end{subfigure}
    
    \begin{subfigure}[b]{0.23\textwidth}
        \centering
        \includegraphics[trim={2cm 0 2cm 0}, clip, width=\textwidth]{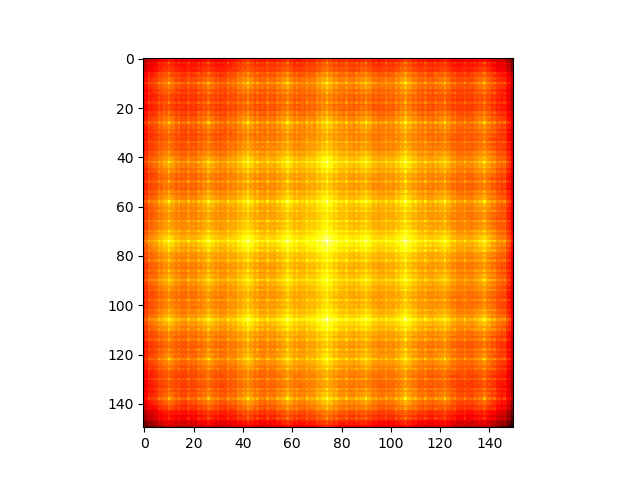}
        \caption{Mobile C}
    \end{subfigure}
    \hfill
    \begin{subfigure}[b]{0.23\textwidth}
        \centering
        \includegraphics[trim={2cm 0 2cm 0}, clip, width=\textwidth]{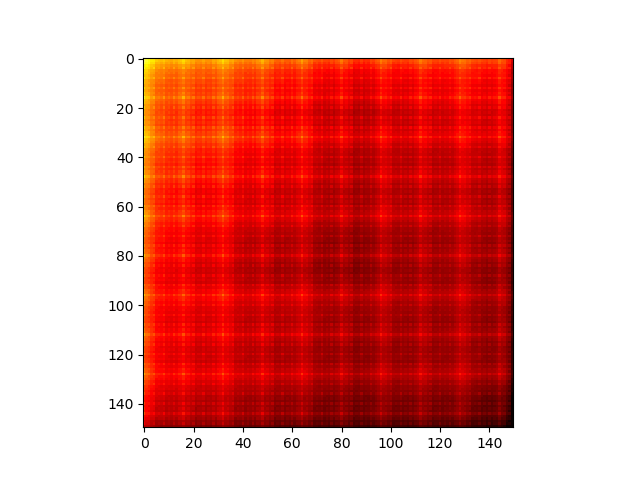}
        \caption{Mobile TL}
    \end{subfigure}
    \hfill
    \begin{subfigure}[b]{0.23\textwidth}
        \centering
        \includegraphics[trim={2cm 0 2cm 0}, clip, width=\textwidth]{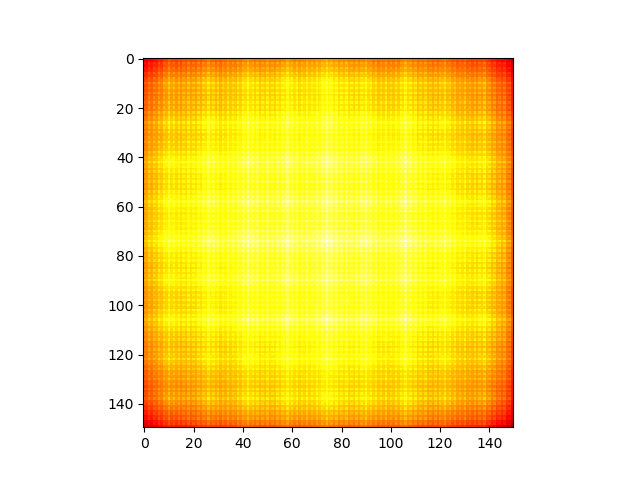}
        \caption{Mobile-AA C}
    \end{subfigure}
    \hfill
    \begin{subfigure}[b]{0.23\textwidth}
        \centering
        \includegraphics[trim={2cm 0 2cm 0}, clip, width=\textwidth]{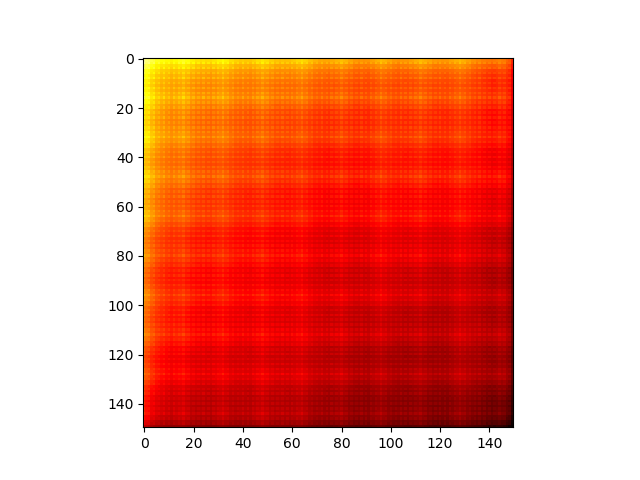}
        \caption{Mobile-AA TL}
    \end{subfigure}
    
    \begin{subfigure}[b]{0.8\textwidth}
        \centering
        \includegraphics[width=\textwidth]{colorbar.PNG}
    \end{subfigure}
    \caption{Cosine similarity heatmaps of the small-patch dataset. Each pixel represents the average cosine similarity of different object patches at that shift location. C indicates the center anchor and TL indicates the top-left anchor.}
    \label{fig:simsmall}
\end{figure}

Similar patterns are visible in the heatmaps for the small-patch dataset in Figure \ref{fig:simsmall} and the aggregate cosine similarities in Table \ref{tab:simsmall}. In fact, these phenomena are even more pronounced, as smaller patches can be shifted further in the image frame. 


\begin{table}[p]
    \centering
    \caption{Feature similarity experiment metrics for the large-patch dataset. $\bar{x}$ is the mean of the mean accuracies and $\sigma$ is the standard deviation of the mean accuracies.}
    \label{tab:simlarge}
    \begin{tabular}{@{} l lll ll ll ll ll @{}}
        \toprule
        \multirow{2}{*}{model} & \multirow{2}{*}{layer} & \multicolumn{2}{c}{top-left} & \multicolumn{2}{c}{top-right} & \multicolumn{2}{c}{center} & \multicolumn{2}{c}{bottom-left} & \multicolumn{2}{c}{bottom-right} \\ \cmidrule(l{2pt}r{2pt}){3-4} \cmidrule(l{2pt}r{2pt}){5-6} \cmidrule(l{2pt}r{2pt}){7-8} \cmidrule(l{2pt}r{2pt}){9-10} \cmidrule(l{2pt}r{2pt}){11-12}
         & & $\bar{x}$ & $\sigma$ & $\bar{x}$ & $\sigma$ & $\bar{x}$ & $\sigma$ & $\bar{x}$ & $\sigma$ & $\bar{x}$ & $\sigma$ \\ \midrule
        \multirow{3}{*}{AlexNet} & fc6 & 0.830 & 0.030 & 0.823 & 0.031 & 0.928 & 0.012 & 0.822 & 0.031 & 0.816 & 0.032 \\
        & fc7 & \textbf{0.949} & 0.016 & \textbf{0.946} & 0.018 & 0.977 & 0.007 & 0.945 & 0.015 & \textbf{0.944} & 0.017 \\
        & fc8 & 0.918 & 0.032 & 0.908 & 0.043 & 0.955 & 0.016 & 0.912 & 0.032 & 0.908 & 0.035 \\
        ResNet-50 & fc & 0.945 & 0.022 & 0.935 & 0.044 & 0.961 & 0.015 & 0.942 & 0.025 & 0.930 & 0.042 \\
        MobileV2 & fc & 0.931 & 0.029 & 0.920 & 0.040 & 0.954 & 0.017 & 0.930 & 0.027 & 0.917 & 0.041 \\
        \midrule
        \multirow{3}{*}{AlexNet-AA} & fc6 & 0.826 & 0.029 & 0.819 & 0.032 & 0.936 & 0.009 & 0.819 & 0.031 & 0.811 & 0.032 \\
        & fc7 & 0.947 & 0.016 & 0.945 & 0.020 & \textbf{0.981} & 0.006 & 0.944 & 0.016 & 0.941 & 0.019 \\
        & fc8 & 0.930 & 0.028 & 0.916 & 0.051 & 0.970 & 0.010 & 0.925 & 0.029 & 0.911 & 0.045 \\
        ResNet-50-AA & fc & 0.948 & 0.023 & 0.943 & 0.031 & 0.975 & 0.011 & \textbf{0.948} & 0.028 & 0.938 & 0.035 \\
        MobileV2-AA & fc & 0.943 & 0.024 & 0.934 & 0.034 & 0.970 & 0.012 & 0.941 & 0.025 & 0.934 & 0.033 \\
        \bottomrule
    \end{tabular}
\end{table}

\begin{table}[p]
    \centering
    \caption{Feature similarity experiment metrics for the small-patch dataset. $\bar{x}$ is the mean of the mean accuracies and $\sigma$ is the standard deviation of the mean accuracies.}
    \label{tab:simsmall}
    \begin{tabular}{@{} l l ll ll ll ll ll @{}}
        \toprule
        \multirow{2}{*}{model} & \multirow{2}{*}{layer} & \multicolumn{2}{c}{top-left} & \multicolumn{2}{c}{top-right} & \multicolumn{2}{c}{center} & \multicolumn{2}{c}{bottom-left} & \multicolumn{2}{c}{bottom-right} \\ \cmidrule(l{2pt}r{2pt}){3-4} \cmidrule(l{2pt}r{2pt}){5-6} \cmidrule(l{2pt}r{2pt}){7-8} \cmidrule(l{2pt}r{2pt}){9-10} \cmidrule(l{2pt}r{2pt}){11-12}
         & & $\bar{x}$ & $\sigma$ & $\bar{x}$ & $\sigma$ & $\bar{x}$ & $\sigma$ & $\bar{x}$ & $\sigma$ & $\bar{x}$ & $\sigma$ \\ \midrule
        \multirow{3}{*}{AlexNet} & fc6 & 0.754 & 0.046 & 0.744 & 0.046 & 0.874 & 0.021 & 0.737 & 0.048 & 0.731 & 0.049 \\
        & fc7 & \textbf{0.931} & 0.015 & \textbf{0.928} & 0.018 & 0.967 & 0.007 & \textbf{0.923} & 0.017 & \textbf{0.923} & 0.019 \\
        & fc8 & 0.889 & 0.039 & 0.877 & 0.053 & 0.937 & 0.022 & 0.865 & 0.051 & 0.863 & 0.058 \\
        ResNet-50 & fc & 0.921 & 0.031 & 0.908 & 0.050 & 0.949 & 0.019 & 0.913 & 0.039 & 0.892 & 0.063 \\
        MobileV2 & fc & 0.897 & 0.043 & 0.883 & 0.058 & 0.935 & 0.026 & 0.882 & 0.047 & 0.868 & 0.066 \\
        \midrule
        \multirow{3}{*}{AlexNet-AA} & fc6 & 0.740 & 0.043 & 0.733 & 0.042 & 0.882 & 0.017 & 0.724 & 0.046 & 0.720 & 0.044 \\
        & fc7 & 0.927 & 0.017 & 0.926 & 0.019 & \textbf{0.971} & 0.006 & 0.921 & 0.018 & 0.920 & 0.019 \\
        & fc8 & 0.898 & 0.039 & 0.885 & 0.067 & 0.956 & 0.013 & 0.879 & 0.051 & 0.877 & 0.058 \\
        ResNet-50-AA & fc & 0.909 & 0.040 & 0.906 & 0.052 & 0.962 & 0.015 & 0.906 & 0.046 & 0.900 & 0.052 \\
        MobileV2-AA & fc & 0.908 & 0.033 & 0.900 & 0.044 & 0.956 & 0.016 & 0.904 & 0.039 & 0.898 & 0.042 \\
        \bottomrule
    \end{tabular}
\end{table}

\subsection{Feature Separability} \label{sec:res-sep}

Next, we investigate the separability of extracted features. After we perform 5-fold cross validation for each classifier on each object patch, we report the mean of the mean accuracies and the mean of the standard deviations for the cross validations. These metrics are reported for both the left-right classifier and the top-bottom classifier. Table \ref{tab:seplarge} reports the metrics for the large-patch dataset while Table \ref{tab:sepsmall} reports the metrics for the small-patch dataset.

The results clearly show that extracted features are well-correlated with patch location. AlexNet's fc6 layer, for example, had a 99.9\% accuracy for left-right classification. However, the top-bottom classifiers show a significantly lower accuracy by about 10\%, as well as a larger standard deviation in cross validation. This seems to conflict with our observations in feature similarity, in which horizontally shifted images had more similar features. Intuitively, more similar features should be less separable.

We propose that features extracted from horizontally shifted patches are more similar and more correlated. While less values differ between horizontally shifted patches, the values that do differ are more meaningful. In contrast, features extracted from vertically shifted patches are less similar and less correlated. This explains the results of both feature similarity and separability experiments.

\begin{table}[t]
    \centering
    \caption{Feature separability experiment metrics for the large-patch dataset. $\overline{Acc}$ is the mean of the mean accuracies and $\bar{\sigma}$ is the mean of the standard deviations of cross validations.}
    \label{tab:seplarge}
    \begin{tabular}{@{} l l ll ll @{}}
        \toprule
        \multirow{2}{*}{model} & \multirow{2}{*}{layer} & \multicolumn{2}{c}{left-right} & \multicolumn{2}{c}{top-bottom} \\ \cmidrule(l{2pt}r{2pt}){3-4} \cmidrule(l{2pt}r{2pt}){5-6}
         & & $\overline{Acc}$ & $\bar{\sigma}$ & $\overline{Acc}$ & $\bar{\sigma}$ \\ \midrule
        \multirow{3}{*}{AlexNet} & fc6 & 0.999 & 0.002 & 0.898 & 0.126 \\
        & fc7 & 0.995 & 0.009 & 0.896 & 0.128 \\
        & fc8 & 0.984 & 0.023 & 0.896 & 0.129 \\
        ResNet-50 & fc & 0.992 & 0.012 & 0.889 & 0.118 \\
        MobileV2 & fc & 0.989 & 0.015 & \textbf{0.884} & 0.119\\
        \midrule
        \multirow{3}{*}{AlexNet-AA} & fc6 & 0.998 & 0.004 & 0.898 & 0.125 \\
        & fc7 & 0.988 & 0.020 & 0.899 & 0.124 \\
        & fc8 & \textbf{0.976} & 0.036 & 0.899 & 0.124 \\
        ResNet-50-AA & fc & 0.990 & 0.016 & 0.893 & 0.124\\
        MobileV2-AA & fc & 0.989 & 0.017 & 0.891 & 0.119\\
        \bottomrule
    \end{tabular}
\end{table}

\begin{table}[t]
    \centering
    \caption{Feature separability experiment metrics for the small-patch dataset. $\overline{Acc}$ is the mean of the mean accuracies and $\bar{\sigma}$ is the mean of the standard deviations of cross validations.}
    \label{tab:sepsmall}
    \begin{tabular}{@{} l l ll ll @{}}
        \toprule
        \multirow{2}{*}{model} & \multirow{2}{*}{layer} & \multicolumn{2}{c}{left-right} & \multicolumn{2}{c}{top-bottom} \\ \cmidrule(l{2pt}r{2pt}){3-4} \cmidrule(l{2pt}r{2pt}){5-6}
         & & $\overline{Acc}$ & $\bar{\sigma}$ & $\overline{Acc}$ & $\bar{\sigma}$ \\ \midrule
        \multirow{3}{*}{AlexNet} & fc6 & 0.997 & 0.004 & 0.897 & 0.127 \\
        & fc7 & 0.985 & 0.022 & 0.896 & 0.129 \\
        & fc8 & 0.966 & 0.043 & 0.895 & 0.130 \\
        ResNet-50 & fc & 0.980 & 0.029 & \textbf{0.893} & 0.124 \\
        MobileV2 & fc & 0.984 & 0.022 & 0.899 & 0.121\\
        \midrule
        \multirow{3}{*}{AlexNet-AA} & fc6 & 0.995 & 0.008 & 0.900 & 0.123 \\
        & fc7 & 0.977 & 0.034 & 0.899 & 0.124 \\
        & fc8 & \textbf{0.959} & 0.056 & 0.898 & 0.127 \\
        ResNet-50-AA & fc & 0.980 & 0.029 & 0.897 & 0.128\\
        MobileV2-AA & fc & 0.986 & 0.021 & 0.899 & 0.123\\
        \bottomrule
    \end{tabular}
\end{table}

\subsection{Feature Arithmetic} \label{sec:res-arith}

\begin{figure}[t]
    \centering
    \begin{subfigure}[b]{0.47\textwidth}
        \centering
        \includegraphics[width=0.95\textwidth]{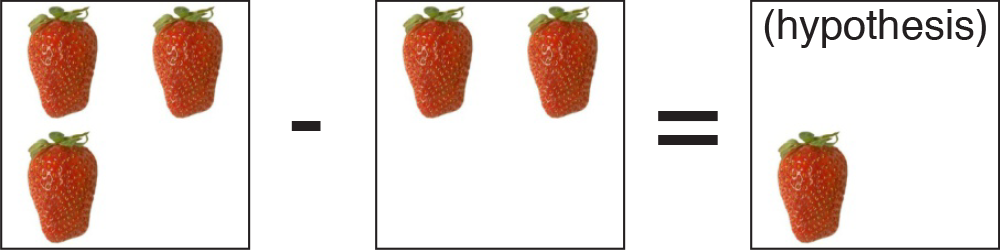}
        \caption{Pixel-space subtraction} \label{fig:6a}
    \end{subfigure}
    \hfill
    \begin{subfigure}[b]{0.47\textwidth}
        \centering
        \includegraphics[width=0.95\textwidth]{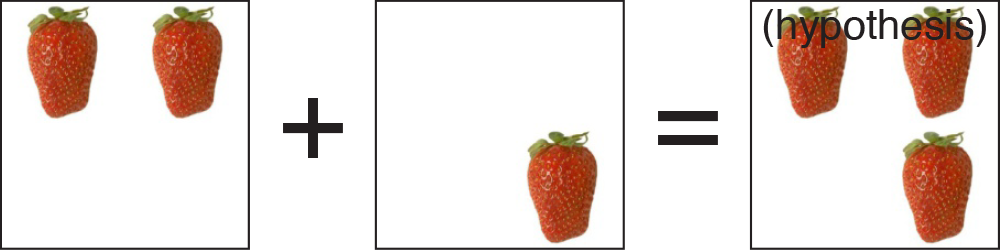}
        \caption{Pixel-space addition} \label{fig:6b}
    \end{subfigure}
    
    \begin{subfigure}[b]{0.47\textwidth}
        \centering
        \includegraphics[width=0.95\textwidth]{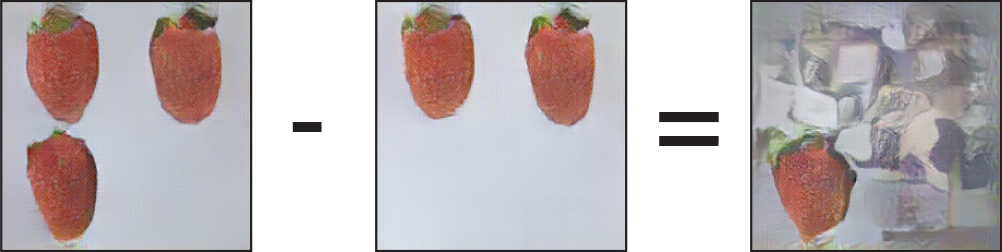}
        \caption{AlexNet fc6 subtraction} \label{fig:6c}
    \end{subfigure}
    \hfill
    \begin{subfigure}[b]{0.47\textwidth}
        \centering
        \includegraphics[width=0.95\textwidth]{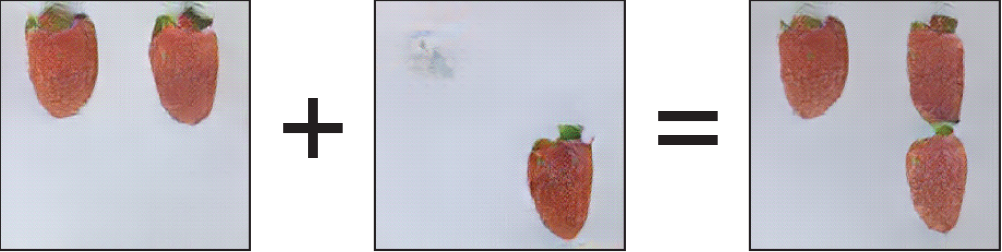}
        \caption{AlexNet fc6 addition} \label{fig:6d}
    \end{subfigure}

    \begin{subfigure}[b]{0.47\textwidth}
        \centering
        \includegraphics[width=0.95\textwidth]{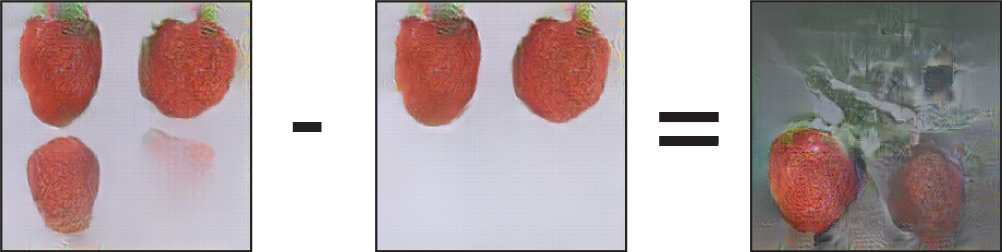}
        \caption{AlexNet fc7 subtraction} \label{fig:6e}
    \end{subfigure}
    \hfill
    \begin{subfigure}[b]{0.47\textwidth}
        \centering
        \includegraphics[width=0.95\textwidth]{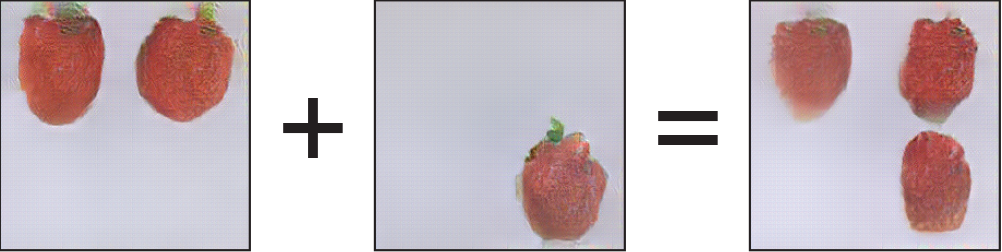}
        \caption{AlexNet fc7 addition} \label{fig:6f}
    \end{subfigure}
    
    \begin{subfigure}[b]{0.47\textwidth}
        \centering
        \includegraphics[width=0.95\textwidth]{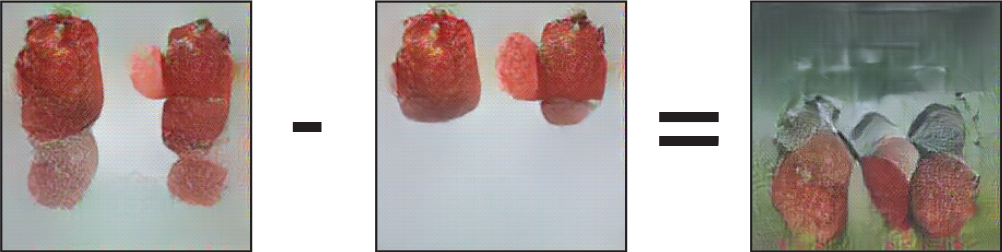}
        \caption{AlexNet fc8 subtraction} \label{fig:6g}
    \end{subfigure}
    \hfill
    \begin{subfigure}[b]{0.47\textwidth}
        \centering
        \includegraphics[width=0.95\textwidth]{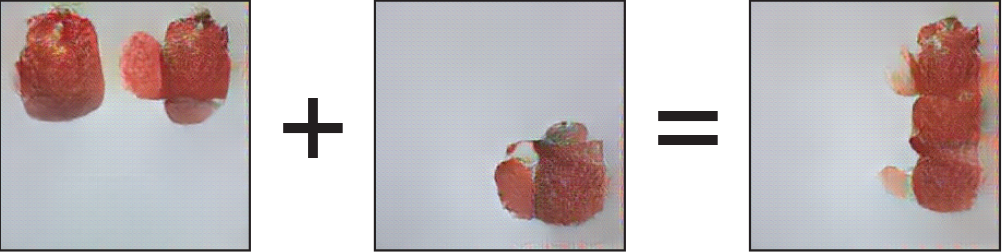}
        \caption{AlexNet fc8 addition} \label{fig:6h}
    \end{subfigure}
    \caption{Feature arithmetic visualizations}
    \label{fig:arith}
\end{figure}

Finally, we report the results of the feature arithmetic experiments. We performed feature arithmetic with several different image patches and operands, but we have chosen the examples in Figure \ref{fig:arith} for discussion. Figure \ref{fig:6a} shows an example of a pixel-space subtraction, where we expect some strawberries in an image to be removed via the subtraction. To perform this extraction in feature space, we extract the features of the two operand images from AlexNet's fc6, fc7, and fc8 layers. Then, for each layer, we calculate a result vector by subtracting the operands' feature vectors. Finally, using DeePSiM \cite{dosovitskiy:deepsim}, we visualize the operand feature vectors and the result vector for each layer. The results are shown in Figures \ref{fig:6c}, \ref{fig:6e}, and \ref{fig:6g}.

Since DeePSiM was trained to reconstruct and visualize extracted feature vectors, the visualizations of the operand feature vectors were expected to be close to the original image. However, DeePSiM was not trained to visualize modified feature vectors, so the visualizations of the result vectors are of interest. For the fc6 layer, despite the noisy background, the visualization of the result vector is similar to the expected image: the top two strawberries have been removed, and only a single strawberry remains in the lower left corner.

For the fc7 and fc8 layers, however, the results are less clear. The visualizations for both result vectors contain an extra strawberry in the bottom right corner, which did not exist in any of the original images. In fact, for fc8, the visualization of the extracted operand features is not accurate, displaying four (or more) strawberries when there were only three originally. Features at these deeper layers seem to encode less spatial information, which is consistent with the results of previous experiments.

Figure \ref{fig:6b} shows an intuitive pixel-space addition. Similarly to the subtraction example, Figures \ref{fig:6d}, \ref{fig:6f}, and \ref{fig:6h} show results from visualizing arithmetic with features from fc6, fc7, and fc8, respectively. In this example, the visualizations of the result vectors from layers fc6 and fc7 closely match the expected image. There are clearly three strawberries in the result, two on the top and one on the bottom right. However, in the visualization of the result vector from layer fc8, the top left strawberry completely disappears, despite the correct visualization of the first operand. As observed in the subtraction experiment, this also suggests that deeper layers encode less spatial information.

In several more experiments with various objects and operand combinations, the above observations remained consistent. Visualizations of the result vector for layer fc6 consistently matched the expected pixel-space results, whereas such visualizations for layers fc7 and fc8 were less consistent or completely inaccurate. This supports that deeper layers encode less spatial information, although some amount can still be recovered due to global invariance.

\section{Conclusion}

Feature extraction with pre-trained convolutional neural networks is a powerful computer vision tool. However, some tasks may require that these extracted features be shift invariant, to be more robust to camera movement or other geometric perturbations. In this work, we measure the shift invariance of extracted features from popular CNNs with three simple experiments and visualizations.

Heatmaps of cosine similarities between extracted features of shifted images reveal that, while various models and layers have different degrees of invariance, none are globally shift invariant. Patterns in the heatmaps show a bias of extracted features towards horizontally shifted images, which remain more similar than vertically shifted images. A grid pattern of local fluctuations is visible in all off-the-shelf models, but anti-aliasing is able to significantly suppress this effect and improve local shift invariance. However, anti-aliasing does not significantly improve global shift invariance.

Results of the features separability experiments suggest that, while features of horizontally shifted images are more similar, they are also more correlated with the geometric transformation. Linear SVM classifiers were significantly better at classifying shifts between left and right than top and bottom. Features of vertically shifted images are more different, but the differences are less correlated and less separable.

Finally, features extracted from some layers of AlexNet can be added and subtracted, and spatial information can still be visualized and recovered. Visualizations of added and subtracted extracted features match the expected pixel-space result for earlier fully connected layers of AlexNet, such as fc6 and fc7. Features extracted from the deepest layer, fc8, are not added and subtracted as easily, as visualizations often do not match the expected pixel-space result.

Further work would be even more valuable to examine the robustness and sensitivity of extracted features. Larger scale experiments with more patch sizes and different background colors and textures may provide better understanding. A more thorough investigation of the separability of extracted features is needed, perhaps from a metric learning approach. Finally, applying the lessons learned to improve the training or architecture of the models is crucial for improving shift invariance, robustness to geometric perturbations, and transferability of models to other tasks.

\clearpage
%
%
\bibliographystyle{splncs04}
\bibliography{egbib}
\end{document}